\documentclass[conference]{IEEEtran}
\IEEEoverridecommandlockouts
\usepackage{cite}
\usepackage{amsmath,amssymb,amsfonts}
\usepackage{algorithmic}
\usepackage{graphicx}
\usepackage{textcomp}
\usepackage{xcolor}
\usepackage{algorithm}
\usepackage{svg}
\usepackage{makecell}
\usepackage{multirow}
\usepackage{caption}

\def\BibTeX{{\rm B\kern-.05em{\sc i\kern-.025em b}\kern-.08em
    T\kern-.1667em\lower.7ex\hbox{E}\kern-.125emX}}
\begin{document}

\title{PR-CIM: a Variation-Aware Binary-Neural-Network Framework for Process-Resilient Computation-in-memory}

\author{
  Minh-Son Le\\
  \texttt{sonlm@khu.ac.kr}
  \and
  Thi-Nhan Pham\\
  \texttt{nhanpt@khu.ac.kr}
  \and
  Thanh-Dat Nguyen\\
  \texttt{datnt@khu.ac.kr}
  \and
  Ik-Joon Chang\\
  \texttt{ichang@khu.ac.kr}
}

\maketitle

\begin{abstract}
Binary neural networks (BNNs) that use 1-bit weights and activations have garnered interest as extreme quantization provides low power dissipation. By implementing BNNs as computing-in-memory (CIM), which computes multiplication and accumulations on memory arrays in an analog fashion, namely analog CIM, we can further improve the energy efficiency to process neural networks. However, analog CIMs suffer from the potential problem that process variation degrades the accuracy of BNNs. Our Monte-Carlo simulations show that in an SRAM-based analog CIM of VGG-9, the classification accuracy of CIFAR-10 is degraded even below 20\% under process variations of 65nm CMOS. To overcome this problem, we present a variation-aware BNN framework. The proposed framework is developed for SRAM-based BNN CIMs since SRAM is most widely used as on-chip memory, however easily extensible to BNN CIMs based on other memories. Our extensive experimental results show that under process variation of 65nm CMOS, our framework significantly improves the CIFAR-10 accuracies of SRAM-based BNN CIMs, from 10\% and 10.1\% to 87.76\% and 77.74\% for VGG-9 and RESNET-18 respectively.
\end{abstract}


\section{Introduction} \label{introduction}
Deep neural networks (DNNs) have shown outstanding performance to surpass human-level accuracy in many applications such as image processing, voice recognition, and language translation. Many researchers have made hard efforts to deploy the inference of DNNs at resource-constraint edge devices, still challenging due to the following reasons. It is well-known that DNNs accompany myriad parameters, and hence, a large memory size is indispensable to operate DNNs efficiently. At resource-constraint edge devices, it is not easy to increase the size of embedded memories above a certain level, leading to many DRAM footprints. In addition, the computational overhead of DNNs is considerable as well. These result in substantial energy dissipation, the major hurdle of edge devices. 

Recently, some researches have shown the possibility to clear the hurdle. One of the most representative researches is a binary neural network (BNN), where all weights and activations are binarized. Despite the extreme 1-bit quantization, BNN delivers reasonably good accuracy \cite{BNN, XNOR, binaryduo}. Due to the aggressive quantization, BNNs have a small model size and can be processed with small-sized embedded memories, ensuring low energy dissipation. The energy efficiency of BNNs is further improved by directly computing BNNs in embedded memories such as SRAM, e-FLASH, and STT-MRAM, namely computation-in-memory (CIM) \cite{BNN/TNN-SRAM, parallelizing_sram, XNOR-free, flash-BNN, MRIMA}. Most BNN CIM hardware computes multiplication and accumulations by using the concept of analog computing, so-called analog CIM, further enhancing energy efficiency. However, in reality, we should consider the potential problem that process variation significantly degrades the accuracy of BNNs to operate on analog CIM platforms. The low-resolution weights of BNNs tend to make BNNs prone to errors due to process variations, strongly motivating techniques to alleviate this problem.    

This work presents a variation-aware BNN framework to deliver accurate analog CIM operations even under process variations of scaled technologies, based on the conventional 6T-SRAM. Recently, many emerging non-volatile memories (eNVM) based CIMs have gathered interest because of their high density and low standby power\cite{Device_Variation_crossArray, TernaryWeight-RRAM, flash-BNN, MRIMA, Fully_parallel_RRAM, XNOR-RRAM}. However, the eNVM-based CIMs face many challenges in manufacturing actual hardware, while SRAM has advantages in terms of the actual design perspective and hence, plays a dominant role in the CIM design. Considering such a situation, we develop the variation-aware BNN framework on the SRAM-based CIM. However, we can easily extend the developed framework for the CIMs of other memories. 

The contribution of this paper can be summarized as follows. 
\begin{itemize}
    \item We derive mathematical models related to the effect of process variations on the SRAM-based BNN CIM circuit. In the CIM circuit, an SRAM cell current represents the multiplication results of the weight and activation corresponding to the SRAM cell. Due to parametric process variations, the SRAM cell current is varied, affecting analog computation. Our model regards such a situation as the weight variation of BNNs. We extracted the models of the weight variations by using Monte-Carlo (MC) simulations in 65nm CMOS.    
    \item Based on the derived model, we present the variation-aware BNN training framework to produce variation-resilient training results. During the training, BNNs are considered bi-polar neural networks due to the weight variations aforementioned. We demonstrate the efficacy of the developed framework through extensive simulations.   
    \item We optimize biasing voltages of bit-lines (BLs) and word-lines (WLs) of SRAM, further improving the accuracy of the SRAM-based CIM.  
\end{itemize}

The remaining part of this paper is organized as follows. In section \ref{sec:pre}, we explain the background regarding BNN, the architecture of SRAM-based CIM, how DNNs can be mapped onto SRAM-based CIM arrays, and in-memory batch normalization. In section \ref{sec:framework}, we present the variation-aware framework and optimization methodology for biasing voltages of BLs and WLs of SRAM. Section \ref{sec:validation} validates the efficacy of our framework. Lastly, we conclude the paper in section \ref{sec:conclusion}.

\section{Preliminaries}\label{sec:pre}
\subsection{Binary Neural Network}\label{sec:bnn}
In BNN, all weights and activations are binarized, significantly enhancing the DNN inference energy efficiency. Many researchers have shown that despite such a low precision format, BNN delivers good inference accuracy \cite{BNN, XNOR, binaryduo}. The firstly introduced BNN \cite{BNN} uses the sign function for the binarization of both weight and activation, where all weights and activations become \textquoteleft+1' or \textquoteleft-1'. However, some state-of-the-art (SOTA) works improved the accuracy of BNN by using the activations of \textquoteleft0' or \textquoteleft1' \cite{binaryduo} while they still employ the sign function for the binarization of weights. Considering such a trend, we use the following activation function. 
\begin{equation}
\label{eq_act_bin}
    BinAct(X) = \begin{cases}
        1, & X \geq thresh \\
        0, & X < thresh  
    \end{cases}
\end{equation}
, where $thresh$ is the activation threshold. Our experiment results, where $thresh$ is assumed as \textquoteleft0.5', are shown in Table \ref{tbl:baseline}, showing the similar trend to SOTA works \cite{binaryduo} as well. Since we use the activation function of (\ref{eq_act_bin}), activations have the values of \textquoteleft0' or \textquoteleft1'. 

\begin{table}[t]
\caption{Comparison of two binary activation cases, (+1/-1) and (1/0)} 
\label{tbl:baseline}
\centering
\resizebox{1.0\linewidth}{!}{
\begin{tabular}{|c|c|c|c|c|c|c|}
\hline
\multicolumn{1}{|c|}{\multirow{2}{*}{Network}} & \multicolumn{1}{c|}{\multirow{2}{*}{Dataset}} & \multicolumn{1}{c|}{\multirow{2}{*}{\shortstack{Full\\precision}}} & \multicolumn{2}{c|}{BNN} & \multicolumn{2}{c|}{{Split BNN}}\\\cline{4-7}
&&&\makecell{(+1/-1)} & \makecell{(1/0)} & \makecell{(+1/-1)} & \makecell{(1/0)}\\
\hline
VGG-9 & CIFAR-10 & 93.71 & 89.77 & 91.36 & 88.51 & 88.70 \\
RESNET-18 & CIFAR-10 & 91.17 & 82.82 & 83.06 & 78.30 & 81.08 \\
\hline
\end{tabular}
}
\end{table}

\subsection{The Architecture of SRAM-based CIM}\label{sec:sram_config}

\begin{figure}[t]
\centering
\includegraphics[width=\linewidth]{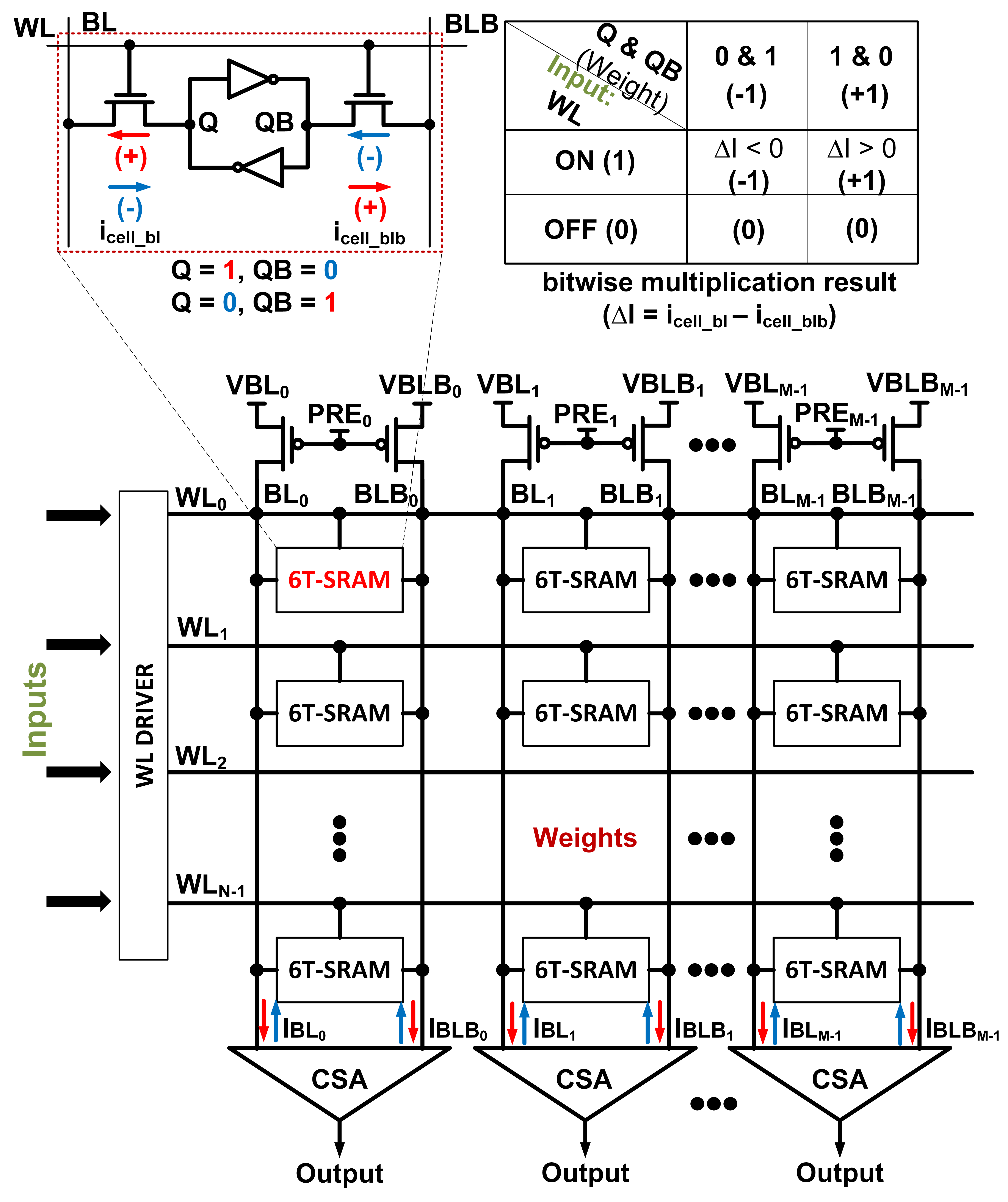}
\caption{6T-SRAM based CIM architecture for a BNN and truth table of input neurons and weights.}
\label{fig:cell}
\end{figure}

\begin{figure}[t]
\centering
\includegraphics[width=0.9\linewidth]{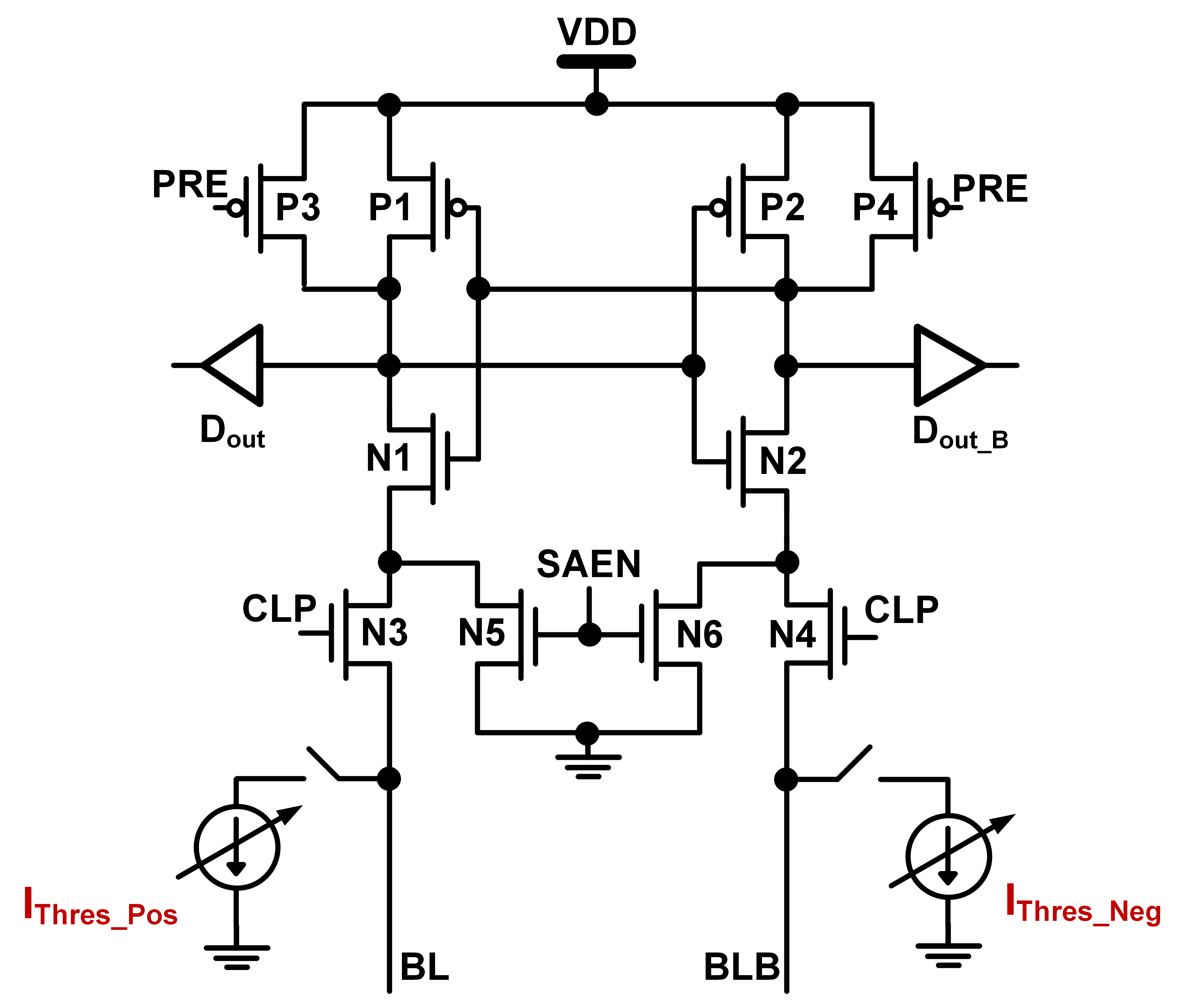}
\caption{Sense amplifier circuit}
\label{fig:SA}
\end{figure}

Fig. \ref{fig:cell} shows the most widely used 6T-SRAM based CIM architecture and cell configuration for the BNN computation, which refers to the design of Rui Liu et al. \cite{parallelizing_sram}. We consider such an architecture for our proposed BNN framework, discussed in section \ref{sec:framework}.

In Fig. \ref{fig:cell}, weights of BNNs are stored to 6T-SRAM cells, and the bitwise multiplications between weights and input activations of the network are directly computed with an analog fashion inside the SRAM array. Let us assume that the weight of \textquoteleft+1' is represented by Q = 1, QB = 0, and the weight of \textquoteleft-1' as the inverted cell. When we operate a BNN on the given configuration, the input activations of a certain BNN layer become the digital values of WLs, since the activation function of (\ref{eq_act_bin}) is considered in this work, as mentioned in section \ref{sec:bnn}. When the inference is executed, all WLs are biased upon the input activations and then, the product of the $i^{th}$ weight and the $i^{th}$ activation becomes the difference between the $bl$ and $blb$ cell currents, \textquoteleft$i_{cell\_bl\_i}$ - $i_{cell\_blb\_i}$' in Fig. \ref{fig:cell}. All cell currents are accumulated to the currents of BL and BLB, implying that \textquoteleft$I_{BL}$-$I_{BLB}$' becomes the multiply-and-accumulation (MAC) output. \textquoteleft$I_{BL}$-$I_{BLB}$' is sensed by the differential current sense-amplifier (CSA), producing the binary activation output based on (\ref{eq_act_bin}). Please, note that when $thresh$ of (\ref{eq_act_bin}) is not zero, a certain circuitry is necessary to implement this. Further, we need to implement batch normalization properly. These are embedded to the sense-amplifier of Fig. \ref{fig:SA}, discussed in section \ref{sec:bn}.

\begin{table}[t]
\caption{Number of split groups for VGG-9} 
\label{tbl:number_of_groups_vgg9}
\centering
\resizebox{1.0\linewidth}{!}{
\begin{tabular}{|c|c|c|}
\hline
\multicolumn{1}{|c|}{\multirow{2}{*}{Layer}} & \multicolumn{1}{c|}{\multirow{2}{*}{Input count per output}} & \multicolumn{1}{c|}{\multirow{1}{*}{Number of groups}} \\
&&\makecell{(Input size = 256)} \\
\hline
1 & 3x3x3 & - \\
2 & 3x3x128 & 6 \\
3 & 3x3x128 & 6 \\
4 & 3x3x256 & 9 \\
5 & 3x3x256 & 9 \\
6 & 3x3x512 & 18 \\
7 & 8192 & 32 \\
8 & 1024 & 4 \\
9 & 1024 & - \\
\hline
\end{tabular}
}
\end{table}

\subsection{Mapping DNNs onto SRAM-based CIM arrays} \label{sec:mappingdnn}

\begin{table}[t]
\caption{Number of split groups for RESNET-18} 
\label{tbl:number_of_groups_resnet18}
\centering
\resizebox{1.0\linewidth}{!}{
\begin{tabular}{|c|c|c|}
\hline
\multicolumn{1}{|c|}{\multirow{2}{*}{Layer}} & \multicolumn{1}{c|}{\multirow{2}{*}{Input count per output}} & \multicolumn{1}{c|}{\multirow{1}{*}{Number of groups}} \\
&&\makecell{(Input size = 256)} \\
\hline
1 & 3x3x3 & - \\
2$\rightarrow$7 & 3x3x16 & 1 \\
8$\rightarrow$13 & 3x3x32 & 2 \\
14$\rightarrow$19 & 3x3x64 & 3 \\
20 & 64 & - \\
\hline
\end{tabular}
}
\end{table}

\subsubsection{Input Splitting}\label{sec:splitting}
\begin{figure}[t] 
\centering
\includegraphics[width=\linewidth]{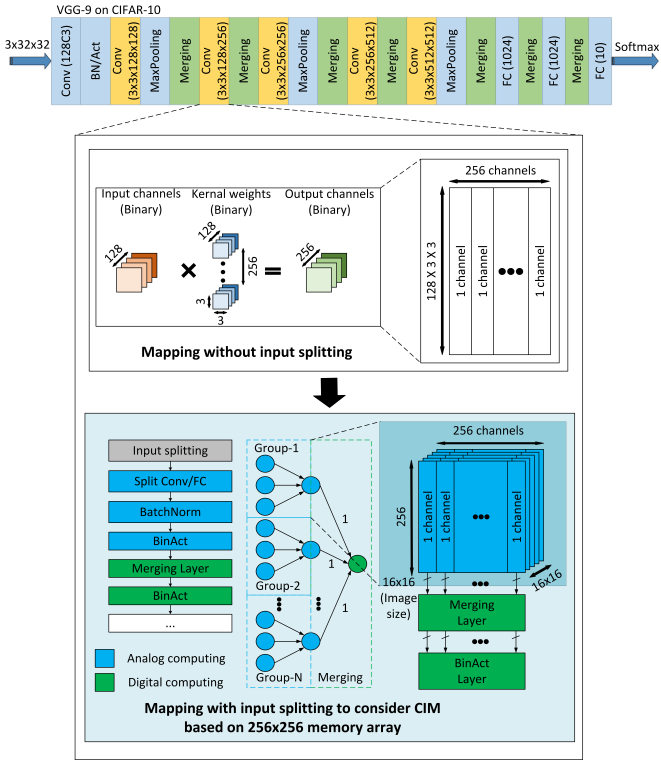}
\caption{SRAM-based CIM mapping}
\label{fig:mapping}
\end{figure}

In the CIM architecture of Fig. \ref{fig:cell}, we store weights to SRAM and control the potentials of WL upon the input activations. Then, SRAM directly computes the matrix multiplications of convolution and fully-connected (FC) layers by using analog computing techniques. In such a scheme, the maximum matrix size that SRAM can calculate at once is dependent on the SRAM array size, which relies on the physical design constraints. Unfortunately, the computed matrix size often exceeds the SRAM array size. Fig. \ref{fig:mapping} shows such a situation well. Here, some convolution layers have 4-dimensional weights. We can regard a convolution layer with 4-dimensional weights as a 2-dimensional matrix with the size of $($kernel size $\times$ kernel size $\times$ input channel size$)$ $\times$ $($output channel size$)$. For instance, in Fig. \ref{fig:mapping}, the 4-dimensional convolution layer whose kernel, input channel, and output channel sizes are 3, 128, and 256, respectively, is considered as the 1152 $(=3 \times 3 \times 128) \times 256$ matrix. To compute the matrix on the circuit of Fig. \ref{fig:cell}, we need 1152 memory rows. Since it is challenging to implement the SRAM array with 1152 rows, we need to properly split the matrix by considering the SRAM array size. Under such a circumstance, an SRAM CIM circuit can deal with one split part of the matrix and produces the corresponding partial sum. All partial sums delivered by the SRAM CIM circuits should be accumulated to complete the matrix computation. SOTA works showed that the precision of the partial sums significantly affects the accuracy of the computed BNNs \cite{parallelizing_sram, XNOR-RRAM, bnn_hardware_codesign, Device_Variation_crossArray}. To obtain multi-bit partial sums in the SRAM CIM circuits, we need ADCs to produce multi-bit outputs, incurring large area and energy overhead. 

To address this problem, the authors of \cite{bnn_hardware_codesign} developed an input splitting technique, which we employed in this work. A large convolution or FC layer is reconstructed into several smaller groups, as shown in Fig. \ref{fig:mapping}, whose input number should be smaller or equal to the number of rows in an SRAM array. Hence, the SRAM array of Fig. \ref{fig:cell} computes the weighted-sums of each group, and the CSAs produce their own 1-bit outputs. Then, the outputs of all groups are merged to fit the input size of the following BNN layer, which is done by digital machines to obtain accurate merging without the effect of process variations. 

The accuracy based on the input splitting technique is compared with the accuracy of the baseline as shown in Table \ref{tbl:baseline}. We assume a typical SRAM array size, $256 \times 256$. The first layer of the BNN that processes the input image and the last layer of the BNN that computes the score of each class are excluded from the input splitting, computed by digital machines. Such an approach follows SOTA works \cite{BNN, XNOR, binaryduo, bnn_hardware_codesign} related to BNN. The split BNN accuracies of Table \ref{tbl:baseline} are considered as the baselines when our techniques are evaluated. For reference, we summarized the number of groups per BNN layer after input splitting on Table \ref{tbl:number_of_groups_vgg9} and Table \ref{tbl:number_of_groups_resnet18} for VGG-9 and RESNET-18 networks respectively.

\subsubsection{Mapping}\label{sec:mapping}
We make more detailed discussions regarding the mapping between convolution layers of BNNs and SRAM-based CIM arrays, shown in Fig. \ref{fig:mapping}. As aforementioned, convolution layers are split to ensure that their size is equal to or less than the number of rows in the SRAM array. As shown in Table \ref{tbl:number_of_groups_vgg9}, the number of split groups of the layer is six $(=\lceil1152/256\rceil+1)$, and the input channels per group are 21 $(=128/6)$ in VGG-9. Hence, the input size of each group is 189 $(=3\times3\times21)$, which is smaller than the number of rows in the SRAM array. Consequently, each group can be regarded as a 2-dimensional matrix with the size of $(3\times3\times21) \times (256)$. Under this circumstance, we can have the mapping strategy that all weights corresponding to each output channel are stored on one column of the SRAM array. The outputs of each group, which is binary (\textquoteleft0' or \textquoteleft1'), are obtained from the macros. We can manage FC layers with the above mapping strategy as well.

\subsection{In-memory batch normalization} \label{sec:bn}

\begin{table}[t]
\centering
\caption{Software to hardware conversion of $BnBinAct()$}
\label{tbl:bn_converted}
\resizebox{1.0\linewidth}{!}{
\begin{tabular}{|c|c|}
\hline
Software implementation & \makecell{Hardware implementation} \\
\hline
{\(BnBinAct(X) = \begin{cases}1, & X \geq X_{th} \\0, & X < X_{th} \end{cases} \)} & \makecell{\( Output(Y) = \begin{cases}1, & Y \geq X_{th} \times \textcolor{red}{IM} \\0, & Y < X_{th} \times \textcolor{red}{IM} \end{cases} \)} \\
& where $Y = I_{BL} - I_{BLB}$ \\
\hline
\end{tabular}
}
\end{table}

\begin{figure}[t]
\centering
\includegraphics[width=0.7\linewidth]{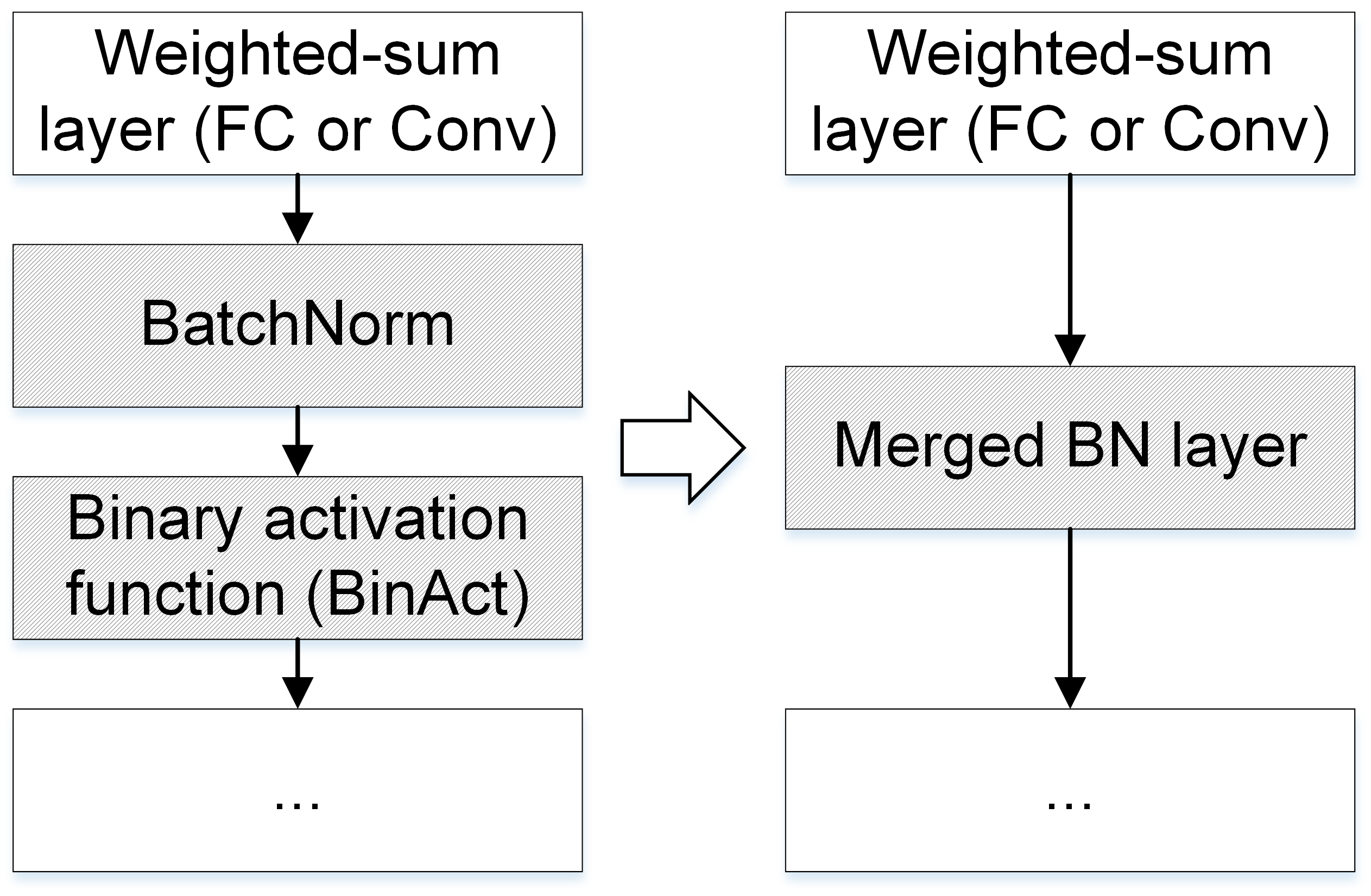}
\caption{Batch normalization merging in inference phase}
\label{fig:bn_merged}
\end{figure}

Batch normalization (BN) is the technique to stabilize the learning process, significantly reducing the number of training epochs. In BNNs, BN highly affects the training accuracy \cite{bn_binary}, more critical. BN can be described by the following equation.
\begin{equation} \label{eq_bn}
    Y = \gamma\frac{X - \mu}{\sqrt{\sigma^2 + \epsilon}} + \beta
\end{equation}
, where X and Y are the input and the output of BN, and $\gamma,\beta,\mu,\sigma$, and $\epsilon$ are a weight, a bias, a mean, a standard deviation, and a sufficiently small constant, respectively. During the back propagation of the training, these four parameters are updated and used to normalize the output of the current batch. In the inference, these parameters become constant, and hence, BN can be regarded as a linear transformation function. As shown in Fig. \ref{fig:bn_merged}, in the inference, the output of the BN layer becomes the input of the activation function (\ref{eq_act_bin}). In this work, we merge (\ref{eq_act_bin}) and (\ref{eq_bn}), whose function is named as $BnBinAct()$. The merged function can be expressed by
\begin{equation}\label{eq_bn_merged}
    BnBinAct(X) = \begin{cases}
        1, & X \geq X_{th} \\
        0, & X < X_{th}  
    \end{cases}
\end{equation}
, where X is the output of weighted-sum layer, and
\begin{equation}
    X_{th} = \frac{(thresh - \beta)}{\gamma}(\sqrt{\sigma^2 + \epsilon}) + \mu.
\end{equation}
Most previous works assume that BN is computed by software \cite{BNN/TNN-SRAM, parallelizing_sram, Fully_parallel_RRAM, XNOR-RRAM}. In such a scenario, ADCs should convert accumulated BL currents to high-precision digital values, which are fed to digital processors, not suitable for edge devices due to large energy overhead. To address this problem, in \cite{batch-merg} BN is implemented as additional cells. In this work, we simply handle the problem by implementing the merged function as the variable current biasing embedded to the differential CSA, shown in Fig. \ref{fig:SA}. 
The biasing current values can be derived from the conversion rule of Table \ref{tbl:bn_converted}. Two current biasing, $I_{Thres\_Neg}$ and $I_{Thres\_Pos}$, are necessary to deal with both positive and negative $X_{th}$ cases. Since the major contribution of this work is to present a variation-aware framework, the detailed discussion regarding the operation of the CSA is not discussed. 

\section{A Variation-aware Binary Neural Network Framework}\label{sec:framework}
\subsection{Variation-aware Models for SRAM-based BNN CIM}\label{sec:model}

\begin{figure}[t]
\centering
\includegraphics[width=0.9\linewidth]{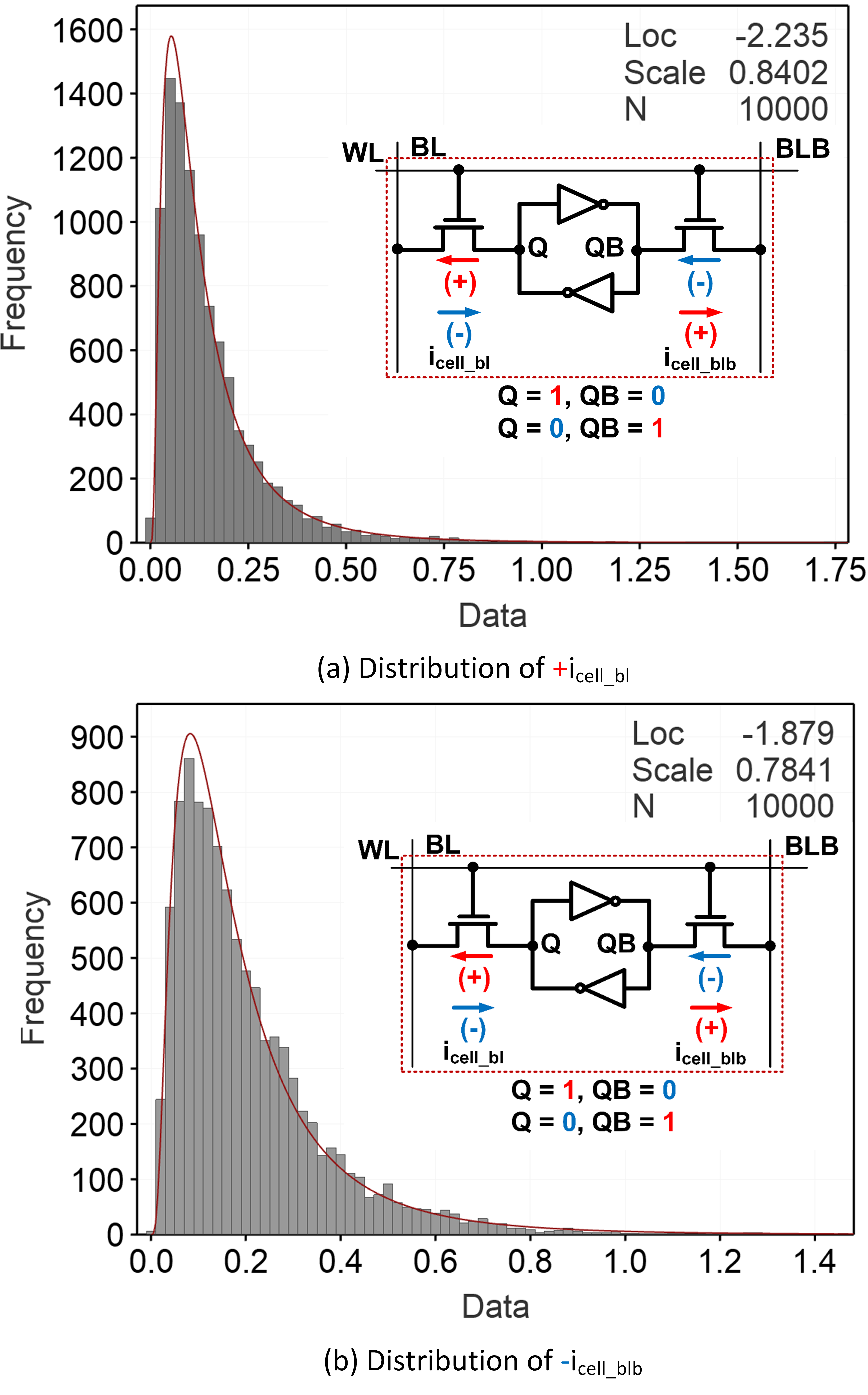}
\caption{Cell current distributions}
\label{fig:cell_dis}
\end{figure}

\begin{figure}[t] 
\centering
\includegraphics[width=\linewidth]{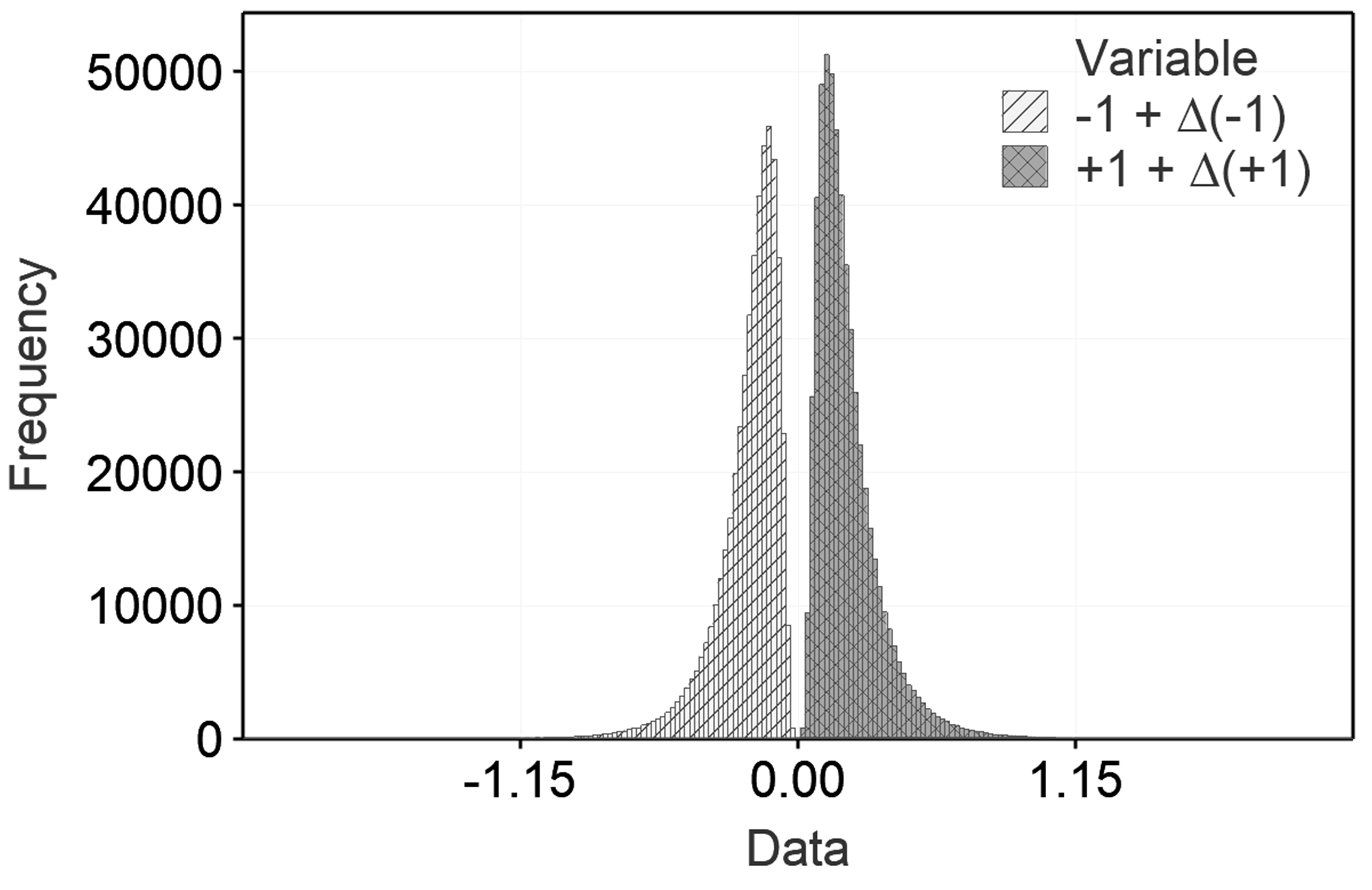}
\caption{Weight distribution under process variations}
\label{fig:weight_dist}
\end{figure}

In this section, we present a variation-aware BNN framework to enhance the reliability of CIM under process variations. The framework assumes SRAM-based CIM, whose configuration is discussed in section \ref{sec:sram_config}. To develop such a BNN framework, firstly, variation-aware models are investigated and derived as follows. 

\begin{figure}[t] 
\centering
\includegraphics[width=\linewidth]{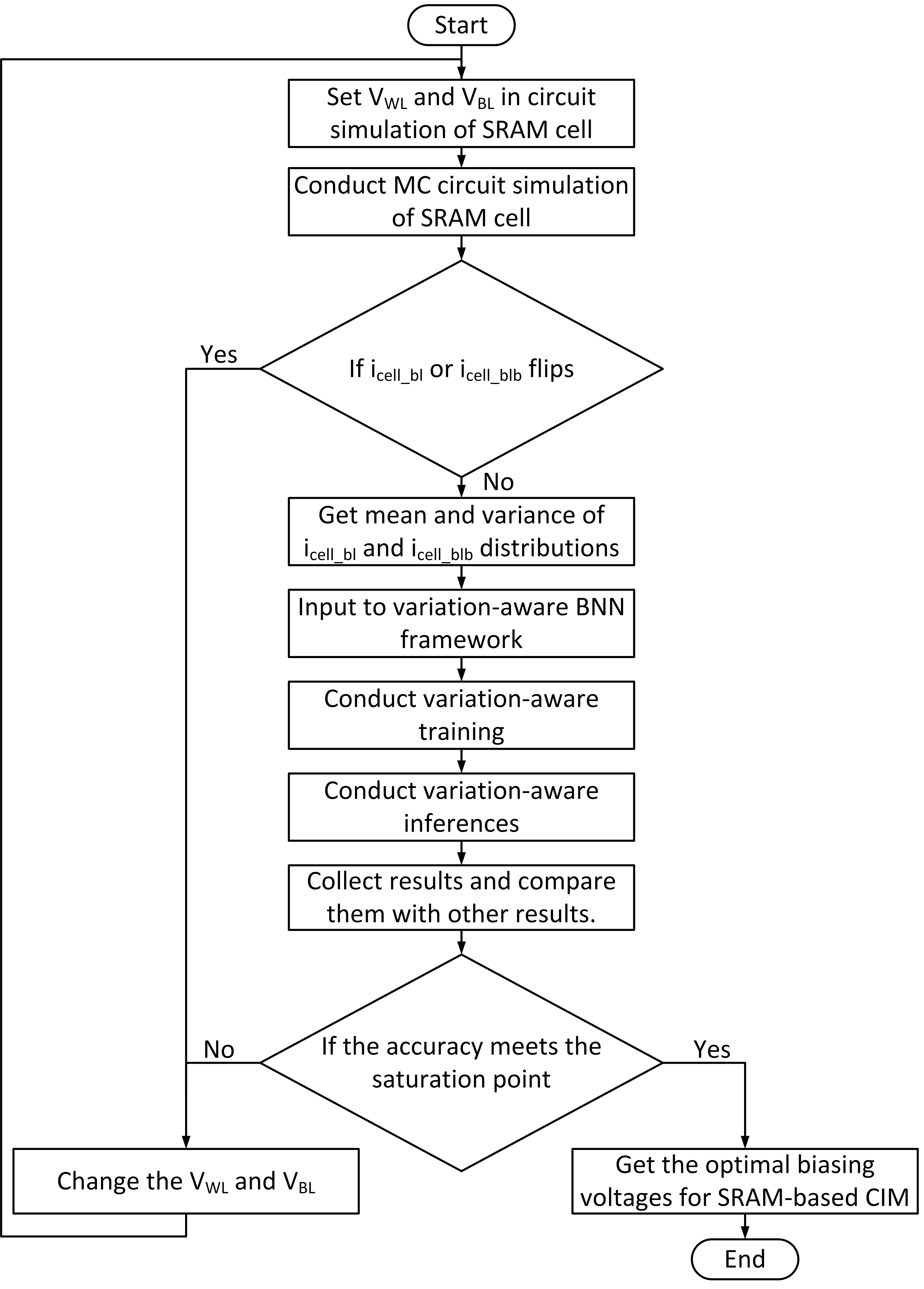}
\caption{Biasing voltages of SRAM optimization methodology}
\label{fig:flowchart}
\end{figure}

In the given configuration (Fig. \ref{fig:cell}), as discussed, the MAC output is defined by \textquoteleft$I_{BL}$-$I_{BLB}$', which described as 
\begin{equation}
\label{eq_total_sram_output}
    \begin{split}
        I_{BL} - I_{BLB} &= \Sigma_{i=0}^{N-1}(i_{cell\_bl\_i} - i_{cell\_blb\_i}) \times WL_{i} \\
        &= \Sigma_{i=0}^{N-1}((W_i) \times IM) \times WL_{i}
    \end{split}
\end{equation}
, where $W_i$ is the $i^{th}$ weight stored in the SRAM array (i.e., $W_i$ is \textquoteleft+1' or \textquoteleft-1'), $WL_i$ is the $i^{th}$ word-line ($WL$) status (ON or OFF), which corresponds to activation values (`1' or `0'), and $IM$ is the current margin that is absolute value of difference current between BL and BLB for one cell (i.e., one bitwise-multiply operation), where no process variations are assumed. In reality, both \textquoteleft$i_{cell\_bl\_i}$' and \textquoteleft $i_{cell\_blb\_i}$' experiences process variations, which can be regarded as the variation of $W_i$ in (\ref{eq_total_sram_output}). Let us model the weight variation as $\Delta_{(W_i)}$. Consequently, the product of the $i^{th}$ weight and the $i^{th}$ activation, \textquoteleft$i_{cell\_bl\_i}$ - $i_{cell\_blb\_i}$', can be redefined as 
\begin{equation}\label{eq_currentvariation}
    \begin{split}
        i_{cell\_bl\_i} - i_{cell\_blb\_i} = (W_i + \Delta_{(W_i)}) \times IM. 
    \end{split}
\end{equation}
In this work, the analysis of process variations was performed through 10,000 MC simulations using statistical models from 65nm CMOS. Without the loss of generality, for one cell configuration as in Fig. \ref{fig:cell} we investigate the case that stored weight as \textquoteleft+1' (i.e., Q = 1 and QB = 0). Therefore, when WL is ON (i.e., input neuron is 1), under process variations, the $bl$ and the $blb$ cell currents have the log-normal distributions of $LN(\mu_{bl}$, $\sigma^2_{bl})$ and $LN(\mu_{blb}, \sigma^2_{blb}$) as shown in Fig. \ref{fig:cell_dis}. Then, we can sequentially derive the following equations. 
\begin{equation}\label{eq_Deltas}
    \Delta_{(W_i)} = \begin{cases}
        \Delta_{(+1)} = \frac{1}{IM} \times $($i_{cell\_bl\_i}$ - $i_{cell\_blb\_i}$) - 1$ \\
        \Delta_{(-1)} = \frac{1}{IM} \times $($i_{cell\_blb\_i}$ - $i_{cell\_bl\_i}$) + 1$
    \end{cases}
\end{equation}

Based on (\ref{eq_Deltas}) and the models that is taken from Fig. \ref{fig:cell_dis}, we obtain the weight distribution under process variations, illustrated in Fig. \ref{fig:weight_dist}. This shows that under process variations of the given CIM configuration, binary weights (-1/+1) are transformed to analog weights ($-1 + \Delta_{(-1)}$/$+1 + \Delta_{(+1)}$) of a BNN, whose distributions are log-normal.

\subsection{Variation-aware Framework for Bi-polar Neural Networks}\label{sec:training}

\begin{algorithm}[t]
\caption{Training a reconstructed L-layer BNN with variation-aware weights and activations. C is the cost function for minibatch, $\lambda$ - the learning rate decay factor and L the number of layers. $\circ$ indicates element-wise multiplication. The function Sign() specifies how to binarize the weights. Polarize() (\ref{eq_polarize}) is used to polarize the binarized one. The activations are clipped to [0, 1] by Clip() function. The function StoQuantize() (\ref{eq_sto_act}) specifies how to binarize the variation-aware activations. BatchNorm() and BackBatchNorm() defines how to batch-normalize and back-propagate the activations, respectively. Update() specifies how to update the parameters when their gradients are known. Straight-Through Estimator (STE) is used for estimating gradients for (\ref{eq_act_bin}) as in \cite{BNN}. Split() and Merge() functions are for input splitting and merging step, as discussed in section \ref{sec:mapping}. $ArraySize$ is the size of SRAM, which is set to 256.}
\label{algo1}
\begin{algorithmic}
\REQUIRE a minibatch of inputs and target $(a_0, a^*)$, previous weights W, previous BatchNorm parameters ($\gamma$, $\beta$), $ArraySize$, weights initialization coefficients from \cite{MSRA_initialization} $\alpha$, and previous learning rate $\eta$.
\ENSURE updated weights $W^{t+1}$, updated BatchNorm parameters ($\gamma^{t+1}$, $\beta^{t+1}$) and updated learning rate $\eta^{t+1}.$ 
\STATE {1. Computing the parameters gradients:}
\STATE {1.1 Forward propagation:}
\FOR{$k = 1$ to L}
    \STATE // Input size per array
    \STATE $InputSize = Kernel \times Kernel \times InputChannels$
    \STATE // Number of groups
    \STATE $nGroups = \lceil InputSize / ArraySize \rceil$
    \WHILE{$InputSize \% nGroups \neq 0$}
        \STATE $nGroups = nGroups + 1$
    \ENDWHILE
    \STATE // Input splitting
    \STATE $a^b_{k-1} \leftarrow Split(a^b_{k-1},nGroups)$
    \STATE $W_k \leftarrow Split(W_k,nGroups) $
    \FOR{$i = 1$ to nGroups}
        \STATE $W_k^b[i] \leftarrow Sign(W_k[i])$
        \STATE $W_k^b[i] \leftarrow Polarize(W_k^b[i])$
        \STATE $s_k[i] \leftarrow a^b_{k-1}[i]W^b_k[i]$
    \ENDFOR
    \STATE $a_k \leftarrow BatchNorm(s_k, \gamma_k, \beta_k)$
    \IF{$k < L$}
        \STATE $a_k \leftarrow Clip(a_k, 0, 1)$
        \STATE $a^b_k \leftarrow StoQuantize(a_k)$
        \STATE $a^b_k \leftarrow Merge(a^b_k,nGroups)$
        \STATE $a^b_k \leftarrow BinAct(a^b_k)$
    \ENDIF
\ENDFOR
\end{algorithmic}
\end{algorithm}

\begin{algorithm}[htpb]
\begin{algorithmic}
\caption*{\textbf{Algorithm} \textbf{\ref{algo1}} [Continued]}
\STATE {1.2 Backward propagation:}
\STATE Compute $g_{a_L} = \frac{\partial C}{\partial a_L}$ knowing $a_L$ and $a^*$
\FOR{$k = L$ to 1}
\IF{$k < L$}
    \STATE $g_{a_k} \leftarrow g_{a^b_k} \circ 1_{0 \leq a_k \leq thresh}$ (STE)
\ENDIF 
    \STATE $(g_{s_k}, g_{\gamma_k}, g_{\beta_k}) \leftarrow BackBatchNorm(g_{a_k}, s_k, \gamma_k, \beta_k)$
    \FOR{$i = 1$ to nGroups}
        \STATE $g_{a^b_{k-1}[i]} \leftarrow g_{s_k[i]}W^b_k[i]$
        \STATE $g_{W^b_k[i]} \leftarrow g^T_{s_k[i]}a^b_{k-1}[i]$
    \ENDFOR
\ENDFOR 
\STATE {2. Accumulating the parameters gradients:}
\FOR{$k = 1$ to L}
    \STATE $\gamma^{t+1}_k \leftarrow Update(\gamma_k, \eta, g_{\gamma_k})$
    \STATE $\beta^{t+1}_k \leftarrow Update(\beta_k, \eta, g_{\beta_k})$
    \FOR{$i = 1$ to nGroups}
        \STATE $W^{t+1}_k[i] \leftarrow Update(W_k[i], \alpha_k[i]\eta, g_{W^b_k[i]})$
    \ENDFOR
    \STATE $\eta^{t+1} \leftarrow \lambda\eta$
\ENDFOR
\end{algorithmic}
\end{algorithm}

The discussion of section \ref{sec:model} shows that with the effect of process variations, each weight stored in memory array experience process variations with the weight variation as $\Delta_{(W_i)}$. Then, the weight stored in each SRAM cell is not an exact digital value of +1 or -1 but can be redefined as 
 
\begin{equation}\label{eq_polarize}
    Polarize(W_i) = \begin{cases}
        +1 + \Delta_{(+1)}, & \text{when } W_i = +1 \\
        -1 + \Delta_{(-1)}, & \text{when } W_i = -1
    \end{cases}
\end{equation}
, where $W_i$ is a binarized weight, and $\Delta_{(+1)}$ and $\Delta_{(-1)}$ are random stochastic parameters to express the effect of process variations, whose distributions are obtained from (\ref{eq_Deltas}). Our training framework is described as Algorithm \ref{algo1}, where the function of (\ref{eq_polarize}) is exploited. In the variation-aware training, we train BNNs based on Algorithm \ref{algo1} from scratch.

Please note that due to process variations of CSA, the activation threshold of (\ref{eq_act_bin}) can be varied. By taking into consideration this, in Algorithm \ref{algo1} we employ the stochastic activation function, whose equation is given by (\ref{eq_sto_act}), instead of the deterministic activation function of (\ref{eq_act_bin}). 
\begin{equation}\label{eq_sto_act}
    StoQuantize(X) = \begin{cases}
        1, & X \geq (thresh + \Delta_{act}) \\
        0, & X < (thresh + \Delta_{act}) 
    \end{cases}
\end{equation}
with 
\begin{equation}\label{eq_delta_act}
    \Delta_{act} \sim N(0, stddev)
\end{equation}
, where the standard deviation of $\Delta_{act}$ is properly assumed. 
When the training step is completed, only binarized weights are left for the inference and the SRAM-based CIM. However, in the inference, the quantized weights need to be polarized again to evaluate the effect of process variation. Since the stochastic function of (\ref{eq_polarize}) is used in the variation-aware inference, we executed the inference 100 times, whose distribution is observed.

\subsection{Optimization of Biasing Voltages}\label{optimization}

In this section, we propose the optimization methodology for biasing voltages of WLs and BLs of SRAM, respectively expressed as $V_{WL}$ and $V_{BL}$, which is shown in Fig. \ref{fig:flowchart}. This methodology provides steps to find the optimal biasing voltages, which delivers the best accuracy. Firstly, the $V_{WL}$ and $V_{BL}$ configuration is set to run MC circuit simulations of the SRAM cell. If a cell flipping occurs, the $V_{WL}$ and $V_{BL}$ configuration is discarded to ensure the accuracy for BNNs. If a cell flipping does not happen, mean and variance of $i_{cell\_bl}$ and $i_{cell\_blb}$ distributions are fed to the variation-aware BNN framework (Section \ref{sec:training}). After conducting the variation-aware training and variation-aware inferences, the average accuracy is collected and compared with those of other $V_{WL}$ and $V_{BL}$ configurations. 

\begin{figure}[t] 
\centering
\includegraphics[width=\linewidth]{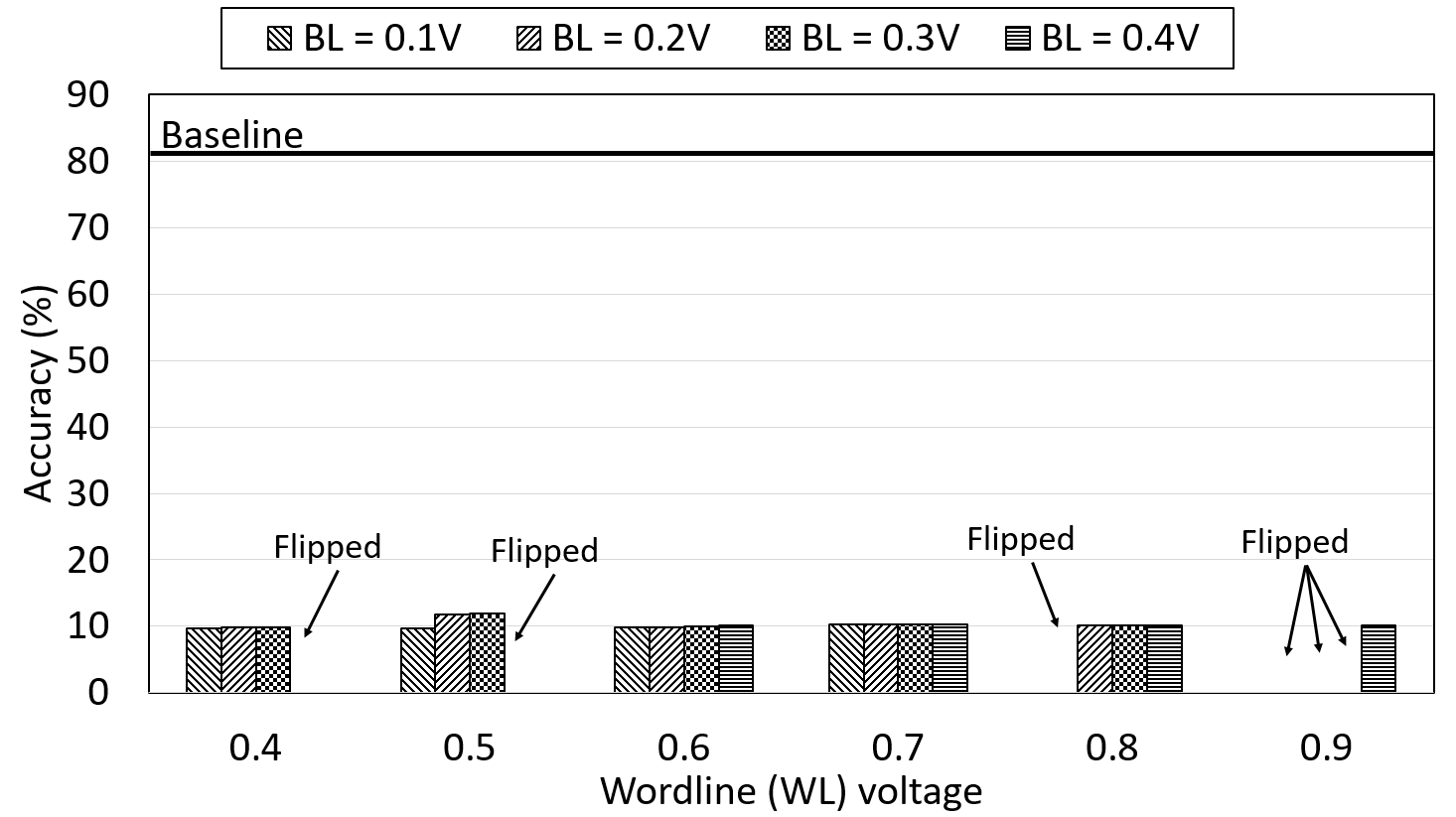}
\caption{Average inference accuracy before variation-aware training of RESNET-18 (full-precision shortcut) on CIFAR-10}
\label{fig:before_resnet18}
\end{figure}

\begin{figure}[t] 
\centering
\includegraphics[width=\linewidth]{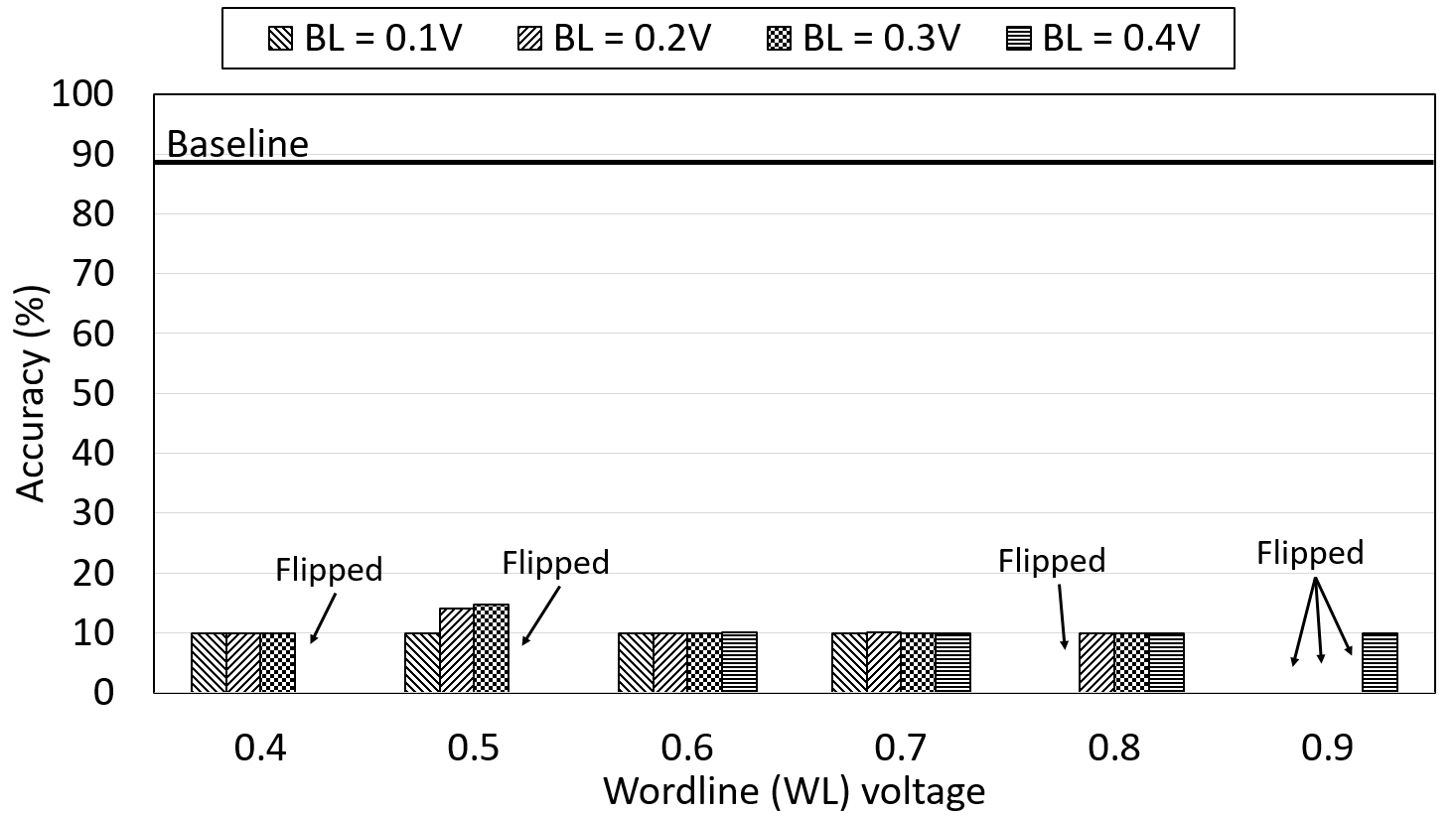}
\caption{Average inference accuracy before variation-aware training of VGG-9 on CIFAR-10}
\label{fig:before_vgg9}
\end{figure}

\begin{figure}[t] 
\centering
\includegraphics[width=\linewidth]{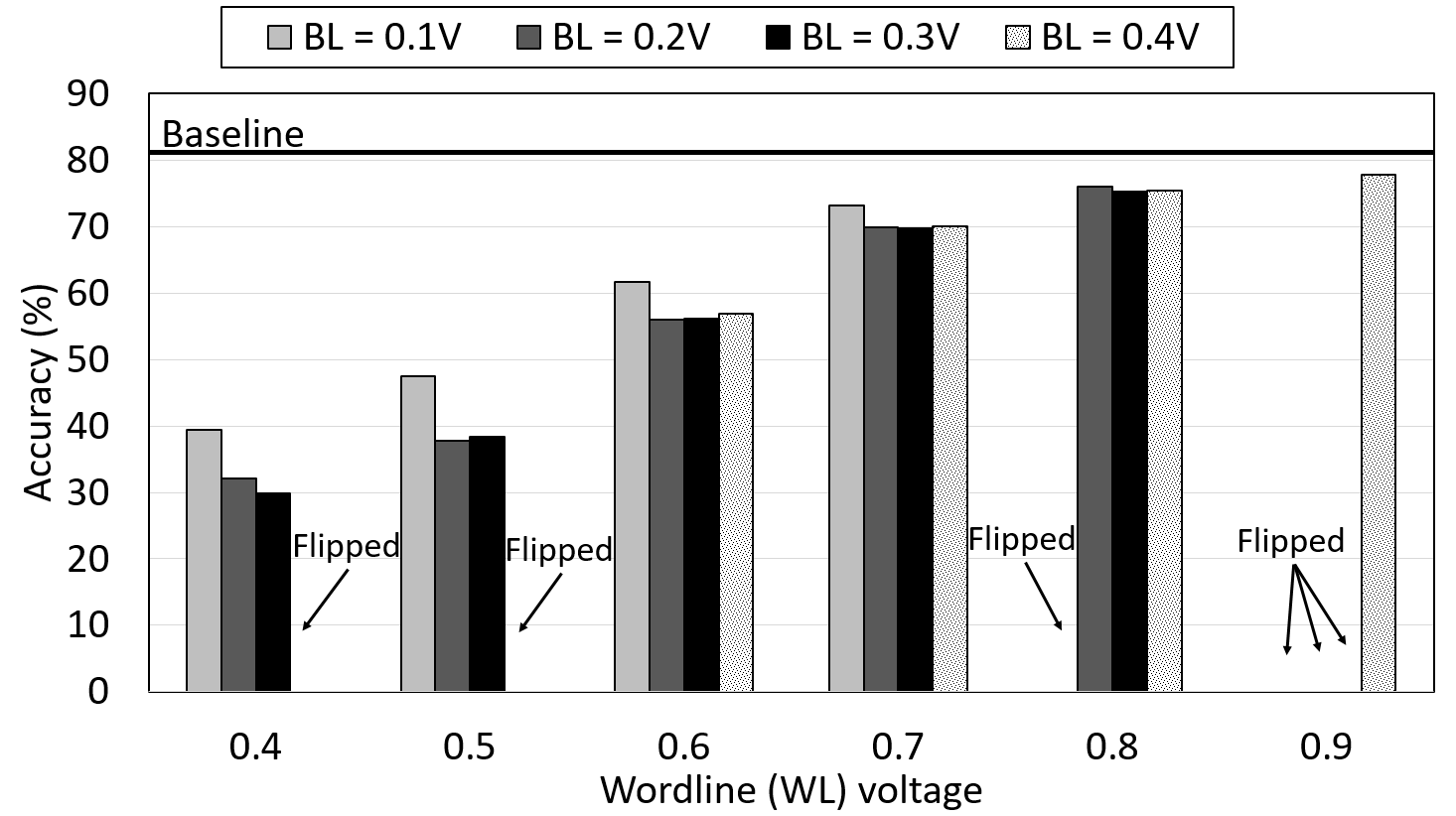}
\caption{Average inference accuracy after variation-aware training of RESNET-18 (full-precision shortcut) on CIFAR-10}
\label{fig:after_resnet18}
\end{figure}

\begin{figure}[t] 
\centering
\includegraphics[width=\linewidth]{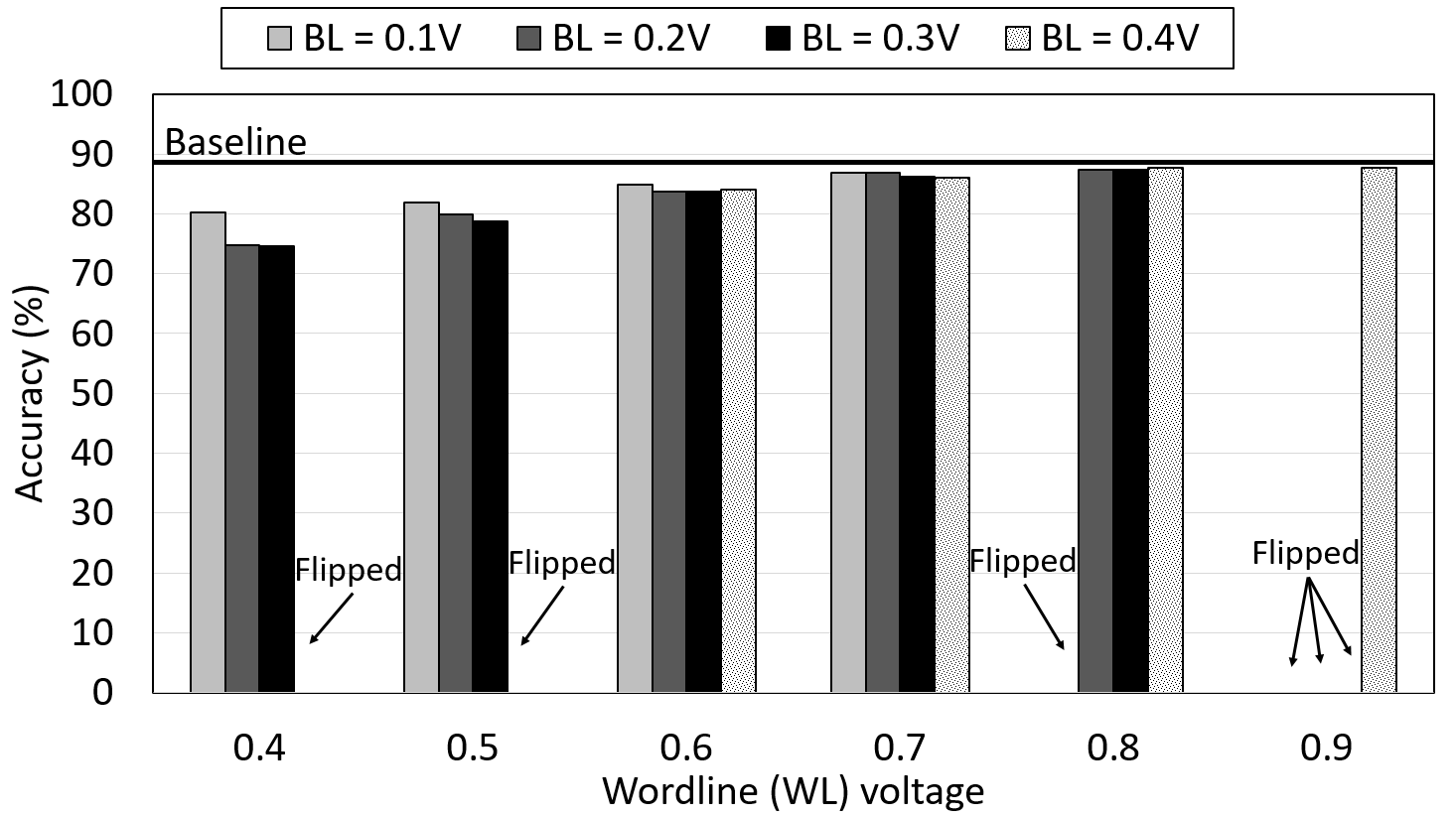}
\caption{Average inference accuracy after variation-aware training of VGG-9 on CIFAR-10}
\label{fig:after_vgg9}
\end{figure}

\subsection{Modeling of IR-drop}\label{sec:ir_drop}
The resistance of the power lines in an SRAM array causes IR-drop, causing the drop of supply voltages. Such an effect is not considered in our experiments. However, we can easily model the drop effect by applying lower supply voltages in our MC simulations.  

\section{Validation of Our Framework}\label{sec:validation}
\subsection{Experimental Setting}\label{sec:setting}
We evaluate the efficacy of our proposed framework with different biasing voltages of BLs and WLs in 6T-SRAM. Under process variations of 65nm CMOS, we estimate the average inference accuracies before and after variation-aware training of RESNET-18 and VGG-9 with CIFAR-10.



In training, the loss is minimized with the algorithm of \cite{adam}, the initial learning rate is set to 0.01, and the maximum number of epochs is 150 for VGG-9 and 250 for RESNET-18. We decay the learning rate by 0.31 when scalar statistics of validation accuracy do not change enough. The variation-aware inferences are 100 times executed with the batch size of 1 for 50000 validation images, whose accuracies are averaged. In (\ref{eq_delta_act}), the standard deviation is assumed as 10\% of $thresh$ in (\ref{eq_act_bin}) and (\ref{eq_sto_act}).  

For the baseline, we do not consider the effect of process variation while the input splitting technique, mentioned in section \ref{sec:mapping}, is used. For RESNET-18, we assume that short-cuts have the data format of full-precision. Many SOTA works \cite{binaryduo} employed such an approach since the accuracy of RESNETs is sensitive to the quantization errors of short-cuts, which is followed in this work. 

\subsection{Results and Discussion}\label{sec:results}
The average inference accuracy before applying variation-aware training of both RESNET-18 and VGG-9, shown in Fig. \ref{fig:before_resnet18} and \ref{fig:before_vgg9}, clearly show that in SRAM-based analog CIMs, the classification accuracy of CIFAR-10 is severely degraded by process variations, even below 20\% for all $V_{WL}$ and $V_{BL}$ voltage configurations under consideration. 

Our variation-aware training framework addresses this problem. Fig. \ref{fig:after_resnet18} shows the inference accuracies of RESNET-18 after using the variation-aware training, where the effect of process variations is considered for the inference as well. The results demonstrate that our variation-aware training framework significantly improves the accuracy under process variations. For instance, when $V_{WL}$=0.9V and $V_{BL}$=0.4V, the accuracy was 10\% under process variations of 65nm CMOS (Fig. \ref{fig:before_resnet18}). Our variation-aware training framework provides a quantum leap for the accuracy of this $V_{WL}$ and $V_{BL}$ biasing condition, to 77.74\%.

We have similar results in VGG-9. Our variation-aware training framework supports remarkable accuracy improvement under the effect of process variations. For instance, in Fig. \ref{fig:before_vgg9}, the accuracies are 10\% for two biasing cases of $V_{WL}$=0.4/$V_{BL}$=0.3, and $V_{WL}$=0.9/$V_{BL}$=0.4 while, in Fig. \ref{fig:after_vgg9}, the accuracies corresponding to the above two cases become 74.51\% and 87.76\%. This well validates the efficacy of our variation-aware training framework.

As illustrated in Fig. \ref{fig:after_resnet18} and \ref{fig:after_vgg9}, the accuracy under process variations tends to increase as $V_{WL}$ increases. It is since cell currents are larger with the higher $V_{WL}$, providing better immunity to process variations. However, when $V_{WL}$ is above a certain level, some SRAM cells are flipped due to a negative read static noise margin \cite{static_noise_margin} (In Fig. \ref{fig:before_resnet18}, \ref{fig:before_vgg9}, \ref{fig:after_resnet18} and \ref{fig:after_vgg9}, the biasing case with the SRAM cell flipping is marked with ``Flipped".). It significantly degrades the accuracy. Considering this factor, we decide the optimal biasing point of $V_{WL}$ and $V_{BL}$, which is the case that $V_{WL}$=0.9V and $V_{BL}$=0.4V. At this point, the CIFAR-10 accuracies of RESNET-18 and VGG-9 are 77.74\% and 87.76\%, respectively. 

\section{Conclusion}\label{sec:conclusion}
By directly computing BNNs in embedded memories, namely computation-in-memory (CIM), we can obtain the utmost energy efficiency for edge devices. However, such an approach suffers from considerable accuracy degradation due to process variation. We present a variation-aware BNN framework on a configuration of SRAM-based CIM, whose efficacy is validated by extensive simulations.

\end{document}